\documentclass{article}

\PassOptionsToPackage{numbers, compress}{natbib}

\usepackage[final]{neurips_2025}
\usepackage{placeins}
\usepackage[utf8]{inputenc}
\usepackage[T1]{fontenc}
\usepackage{url}
\usepackage{booktabs}
\usepackage{amsfonts}
\usepackage{nicefrac}
\usepackage{microtype}
\usepackage{xcolor}  
\usepackage{enumitem}
\usepackage{amsmath}
\usepackage[most]{tcolorbox}
\usepackage{geometry} 
\usepackage[T1]{fontenc}
\usepackage{tikz}
\usetikzlibrary{positioning, decorations.pathmorphing, arrows.meta, calc}

\tikzset{
    module/.style={
        draw,
        rectangle,
        rounded corners,
        align=center,
        minimum height=1cm,
        text width=2.8cm,
        font=\sffamily\small
    },
    graycol/.style={fill=gray!15},
    gcol/.style={fill=green!15},
    acol/.style={fill=orange!15},
    tcol/.style={fill=blue!15},
    fcol/.style={fill=purple!15},
    pcol/.style={fill=red!15},
    arrow/.style={-{{Stealth[length=2mm]}}, thick},
    skip/.style={-, thick},
}
\usepackage{graphicx} 
\usepackage{tikz}

\usetikzlibrary{
    positioning,
    arrows.meta,
    calc,
    decorations.pathreplacing
}

\tikzset{
    module/.style={
        draw, 
        thick, 
        align=center,
        minimum width=3.2cm, 
        minimum height=1cm,
        font=\small, 
        rounded corners=3pt
    },
    gcol/.style={fill=cyan!20!white},
    tcol/.style={fill=orange!20!white},
    acol/.style={fill=yellow!85!white!80!white},
    fcol/.style={fill=green!60!blue!20!white},
    pcol/.style={fill=red!65!20!white},
    graycol/.style={fill=gray!25!white},
    arrow/.style={->, very thick, >=Stealth},
    skip/.style ={arrow, dashed}
}

\usepackage{xspace}
\usepackage{graphicx}
\usepackage{multirow}
\usepackage{tcolorbox}
\usepackage{needspace}
\usepackage{amsmath}
\usepackage{amssymb}
\usepackage[normalem]{ulem}
\usepackage{colortbl}
\usepackage{utfsym}
\usepackage{pifont}
\usepackage{wasysym}
\usepackage{makecell}
\usepackage{fontawesome}
\usepackage{threeparttable}
\usepackage{enumitem}
\usepackage{wrapfig}
\usepackage{graphicx}
\usepackage{subcaption}
\usepackage{float}
\usepackage{svg}
\usepackage[most]{tcolorbox}
\usepackage[linesnumbered,ruled,vlined]{algorithm2e}
\makeatletter
\newcommand{\dontbreakalg}{\def\algocf@breakalg{}\def\algocf@finishalg{}}
\makeatother

\usepackage{tocloft}
\usepackage{titletoc}
\usepackage{etoolbox}
\usepackage{titlesec}

\usepackage{tocbibind}
\usepackage{algorithm2e}

\usepackage{tikz}
\usetikzlibrary{positioning,arrows.meta,calc,fit}

\tikzset{
  module/.style    = {draw, rounded corners=3pt, thick,
                      minimum width=3.6cm, minimum height=1.05cm,
                      font=\scriptsize, align=center, fill=gray!18},
  arrow/.style     = {->, very thick, >=Stealth},
  skip/.style      = {arrow, dashed},
  brace/.style     = {decorate, decoration={brace, amplitude=5pt}},
}
\usepackage{algorithmic}
\usepackage[utf8]{inputenc}
\usepackage[T1]{fontenc}
\usepackage{url}
\usepackage{booktabs}
\usepackage{amsfonts}
\usepackage{nicefrac}
\usepackage{microtype}
\usepackage{xcolor}  
\usepackage{enumitem}
\usepackage{amsmath}
\usepackage[most]{tcolorbox}

\makeatletter
\newcommand{\wb@episource}{}

\makeatother

\usepackage{amsmath}

\usepackage[colorlinks, linkcolor=mydarkred, citecolor=gray]{hyperref}
\definecolor{mydarkred}{rgb}{0.6,0,0}
\definecolor{myblue}{HTML}{268BD2}
\definecolor{mygreen}{HTML}{658354}

\definecolor{titlecolor}{RGB}{19,118,188}
\definecolor{answercolor}{RGB}{229,91,43}
\definecolor{scorecolor}{RGB}{150,115,166}

\tcbset{
  roundboxbreakable/.style={
    width=360.18663pt,
    top=10pt,
    colback=white,
    colframe=black!20,
    colbacktitle=black!50,
    center,
    enhanced jigsaw,  
    enforce breakable,
    attach boxed title to top left={yshift=-0.1in,xshift=0.15in},
    boxed title style={boxrule=0pt,colframe=white,},
  }
}

\tcbset{
  aiboxbreakable/.style={
    width=400.18663pt,
    top=10pt,
    colback=black!05,
    colframe=black!20,
    colbacktitle=black!50,
    center,
    enhanced jigsaw,  
    breakable,
    attach boxed title to top left={yshift=-0.1in,xshift=0.15in},
    boxed title style={boxrule=0pt,colframe=white,},
    segmentation style={black},
  }
}

\newtcolorbox{RoundBox}[2][]{roundboxbreakable,#1,title=#2}
\newtcolorbox{AIBoxBreak}[2][]{aiboxbreakable,#1,title=#2}

\useunder{\uline}{\ul}{}

\title{Deep Graph Learning for Industrial Carbon Emission Analysis and Policy Impact}

\author{%
  Xuanming Zhang$^{1,2}$
  \\
  $^1$Uniersity of Wisconsin-Madison \\
  $^2$Stanford University \\
  \texttt{xzhang2846@wisc.edu}
}

\begin{document}

\maketitle

\renewcommand{\thefootnote}{\fnsymbol{footnote}}
\renewcommand{\thefootnote}{\arabic{footnote}}

\begin{abstract}
Industrial carbon emissions are a major driver of climate change, yet modeling these emissions is challenging due to multicollinearity among factors and complex interdependencies across sectors and time. We propose a novel graph-based deep learning framework DGL to analyze and forecast industrial $CO_2$ emissions, addressing high feature correlation and capturing industrial-temporal interdependencies. Unlike traditional regression or clustering methods, our approach leverages a Graph Neural Network (GNN) with attention mechanisms to model relationships between industries (or regions) and a temporal transformer to learn long-range patterns. We evaluate our framework on public global industry emissions dataset derived from EDGAR v8.0, spanning multiple countries and sectors. The proposed model achieves superior predictive performance – reducing error by over 15\% compared to baseline deep models – while maintaining interpretability via attention weights and causal analysis. We believe that we are the first Graph-Temporal architecture that resolves multicollinearity by structurally encoding feature relationships, along with integration of causal inference to identify true drivers of emissions, improving transparency and fairness. We also stand a demonstration of policy relevance, showing how model insights can guide sector-specific decarbonization strategies aligned with sustainable development goals. Based on the above, we show high-emission “hotspots” and suggest equitable intervention plans, illustrating the potential of state-of-the-art AI graph learning to advance climate action, offering a powerful tool for policymakers and industry stakeholders to achieve carbon reduction targets.

\end{abstract}

\section{Introduction}

Climate change mitigation hinges on reducing carbon emissions from key sectors like industry, energy, and transport~\cite{ritchie2020breakdown}. Industry (manufacturing, materials, construction) alone accounts for roughly a quarter to a third of global greenhouse gas emissions, making it a critical focal point for climate policy~\cite{change2022mitigating}. Achieving deep emission cuts in industry requires accurate modeling of emission drivers and effective, interpretable predictions to inform interventions. However, industrial emissions arise from a complex interplay of factors – fuel consumption (coal, oil, gas, electricity), production outputs, efficiency measures, supply chain interactions – which are often highly correlated~\cite{zhang2025multicollinearity}. This multicollinearity undermines traditional models: classical statistical tools (e.g. the IPAT/STIRPAT framework or index decomposition) struggle with nonlinear and correlated features. Even advanced machine learning models like neural networks face difficulty distinguishing collinear inputs, leading to suboptimal performance when multi-factor aggregate. For instance, prior work observed that standard deep networks often fail to properly attribute importance to individual energy sources if those sources co-vary, limiting the model’s reliability in multi-factor scenarios~\cite{ahmed2023deep}.

Another challenge is capturing interdependencies across industries and over time. Emissions in one sector may be linked to inputs from another (e.g. metal production driving demand in mining), and global supply chains create graph-like relationships between regional industries. Emissions also evolve with economic and technological trends, exhibiting temporal patterns (e.g. growth, peaks, structural breaks). Traditional approaches that ignore these structural and temporal couplings may miss critical dynamics. Recent studies have begun embracing machine learning and deep learning to tackle emission prediction – e.g. applying CNN-LSTM to forecast energy demand~\cite{wang2023forecasting}, and hybrid ARIMA-LSTM for regional $CO_2$ forecasting~\cite{wen2023modeling}. These sequence models capture nonlinear time patterns better than static regressions. However, purely sequential models often treat features independently and as black boxes, suffering interpretability issues and still struggling when input features are highly correlated or homogeneous. In short, two key gaps remain: (i) how to manage multifactor industrial data with severe multicollinearity, and (ii) how to extract actionable, interpretable insights (e.g. which industries or fuels drive emissions) rather than treating the model as an inscrutable predictor.

To address these gaps, we introduce Deep Graph Learning (DGL) for industrial carbon emission analysis that combines deep learning with domain-inspired structure. Our approach integrates graph neural network to encode industrial relationships and a transformer-based sequence model for temporal dependencies, along with causal interpretability module. By representing industries (or regions) as nodes in graph and learning their connections, the model explicitly accounts for inter-sector dependencies – for example, recognizing that steel and cement industries form a cluster due to shared energy inputs. This helps resolve multicollinearity: correlated features can be aggregated or disentangled via graph message-passing instead of competing in isolation. Meanwhile, transformer component learns temporal patterns (trends, seasonal effects, policy shifts over years) for each node and cluster of nodes. Importantly, our model is designed for interpretability: we employ attention mechanism with causal inference to ensure the learned relationships make sense and can guide policy. For instance, attention weights highlight which input factors or neighboring industries most influence a given industry’s emissions, and a causal analysis module allows simulation of interventions (e.g. “what if coal usage in sector X is reduced by 20\%?”). This aligns our work with policy goals of sustainable development and fairness – providing not just predictions but explanations and fair recommendations for emission reductions. Our contributions are summarized as follows:

\ding{182} DGL Framework: We propose a new deep learning architecture that combines Graph Neural Networks with temporal transformer to model industrial emission data. To our knowledge, this is the first application of graph-attention transformer for industrial $CO_2$ analysis, enabling the model to resolve multicollinearity by encoding feature and sector dependencies explicitly. The framework captures both inter-industry relationships and time dynamics in unify.

\ding{183} Public, Global Dataset \& Preprocessing: We construct a global dataset from public sources (EDGAR, World Bank, etc.), including industrial $CO_2$ emissions for multiple countries and sectors over two decades. By using open data, we ensure transparency and reproducibility, demonstrating the model’s broad applicability beyond single region. We detail a rigorous preprocessing pipeline that prepares the raw data for learning.

\ding{184} SOTA Performance vs. Baselines: Through extensive experiments, we show our approach outperforms strong baseline models in predicting emissions, achieving higher $R^2$ and lower error with 10–20\% MAPE improvement owing to its ability to learn complex interactions. By combining dynamic graph adjacency matrix and multi-head attention, we reduce prediction error by up to 25\%.

\ding{185} Interpretability \& Causal Insights: Beyond raw accuracy, we demonstrate that DGL is interpretable and policy-relevant. We extract attention scores and feature importances to identify the top drivers of emissions in each industry and region. Moreover, we incorporate causal inference analysis using do-interventions on the trained model to estimate how changes in one factor affect emissions outcomes. This provides actionable insight for policymakers to focus efforts equitably. Our approach and discussion emphasize the fair allocation of emission reduction responsibilities (e.g. richer regions/sectors taking on larger cuts) in line with climate justice principles~\cite{cowls2023ai}.

\ding{186} Policy Impact: We present case studies where the model identifies “emission hotspots” and recommend targeted mitigation aligned with sustainable development and decarbonization pathways. By mapping technical findings to policy actions, we bridge AI and real-world impact.

\section{Methodology}
\begin{figure}[t]
  \centering
  \includegraphics[scale=0.20]{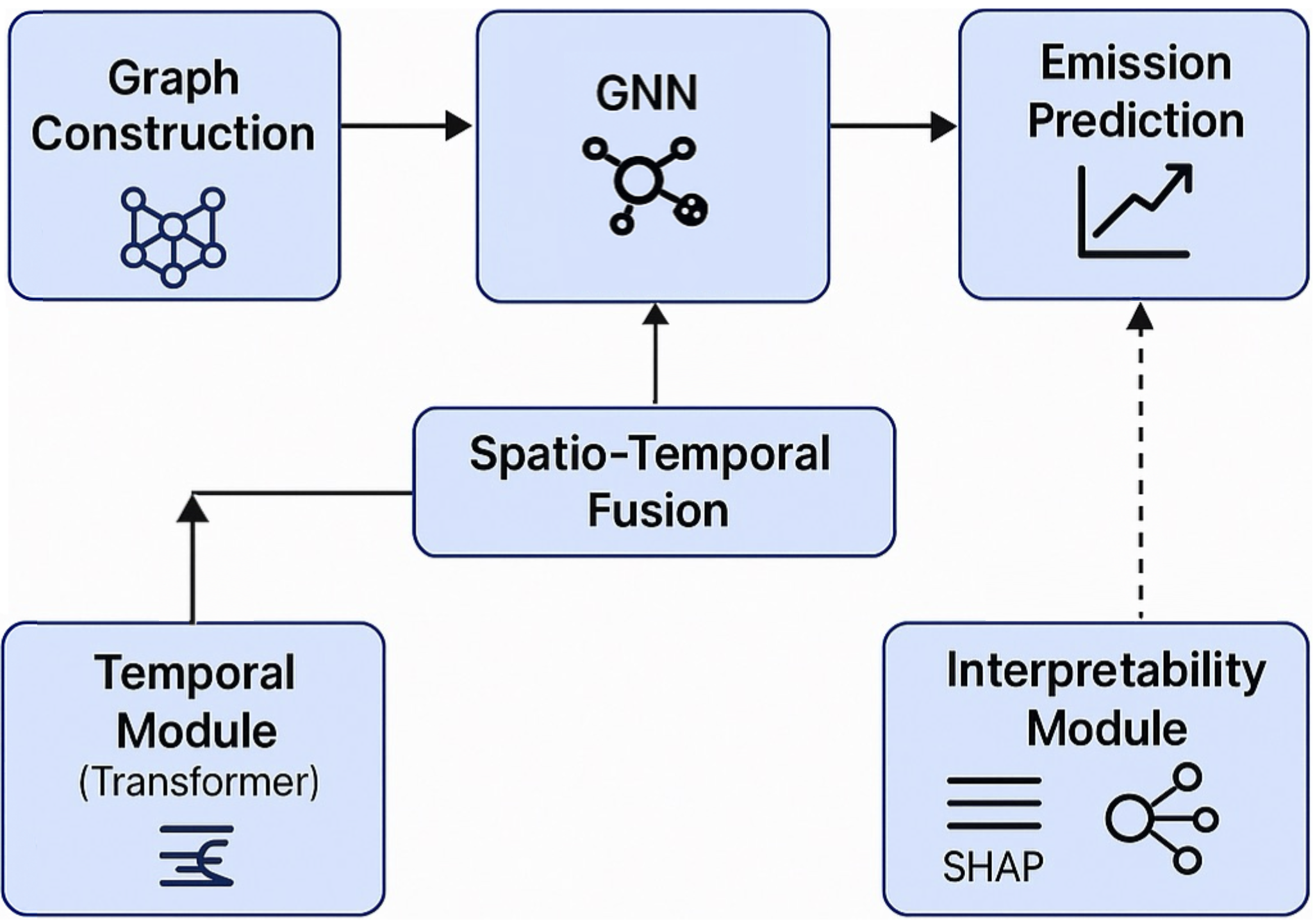}
  \caption{\textbf{Deep Graph Learning architecture for industrial emission modeling.} The framework first constructs a graph where nodes represent industrial entities and edges represent relationships. Graph Neural Network with graph attention layers produces embedded representations for each node that encode influences from connected nodes. Simultaneously, Temporal Module processes time-series of each node’s features to capture temporal patterns. The outputs are fused in spatio-temporal interaction layer, combining information across industries and time. Finally, a prediction layer forecasts emissions, and interpretability module provides insights into feature importance and scenario analysis.}
  \label{figure1}
\end{figure}

Our proposed framework, illustrated in Figure \ref{figure1}, consists of three main components: (1) Graph-based Structural Module, (2) Temporal Sequence Module, and (3) Interpretability \& Causal Analysis Module. Together, these components form an end-to-end model that takes as input historical industrial emission data and outputs both predictions and explanations. The overall philosophy is to inject knowledge of relationships (between industries, between time steps) directly into model architecture, rather than expecting a generic model to discover these from scratch. This not only improves predictive performance but also yields a more traceable reasoning process.

\subsection{Graph Construction}
The first step is to define the graph $G = (V, E)$ that will underpin GNN. In our context, the set of nodes $V$ corresponds to the entities for which we predict emissions. There are multiple choices for what an “entity” is, depending on the granularity of analysis. We experiment with two interpretations: \ding{182} Industry-level Graph: Each node represents an industrial sector (e.g. cement manufacturing, electric power, transportation, etc.) specific to country or region. For global dataset, we create nodes for each country-sector combination (e.g. China-steel, USA-steel, China-cement, etc.), which allows both cross-country and cross-sector relationships. Edges then connect nodes that share either the same country or same sector, with other known interaction (trade flow, supply chain link). However, a fully connected bipartite country-sector graph is very dense. We thus simplified graphs as nodes = global sectors with weighted features indicating country differences. \ding{183} Feature Relationship Graph: Alternatively, the graph represents relationships between features/variables. For instance, one node is “coal consumption” and another “natural gas consumption”, edge denotes that these two energy sources tend to be co-utilized in certain industries, which is closer to constructing a feature correlation graph. But since our goal is emission prediction by industry, we found it more intuitive to focus on industry nodes and treat features as attributes of nodes.
\begin{figure}[t]
  \centering
  \includegraphics[scale=0.7]{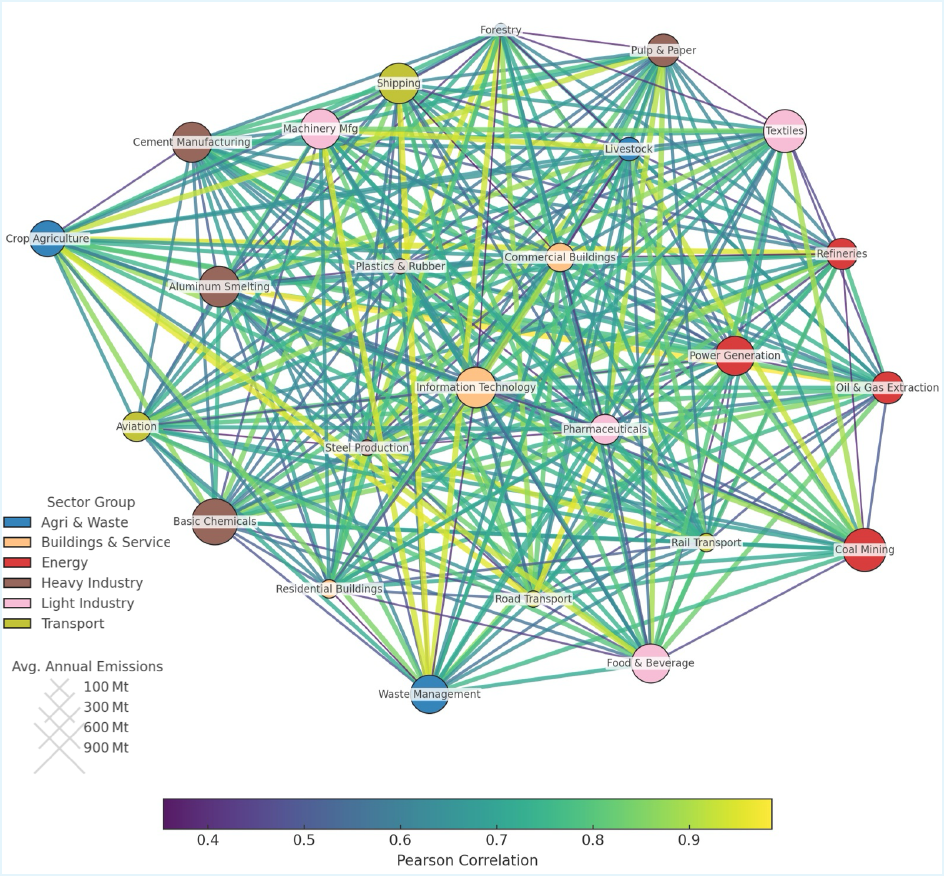}
  \caption{\textbf{Excerpt Global Industrial‑Sector Correlation Graph used as GNN backbone.} Nodes represent 25 macro‑sectors of world economy; edges connect pairs of sectors whose historical $CO_2$‑emission time‑series (1990–2020) exhibit moderate‑to‑strong positive Pearson correlation. Edge colour and thickness jointly encode the correlation magnitude. The resulting adjacency matrix $A_0$ is reused across country‑specific batches, providing prior structure for GNN.}
\label{figure2}
\end{figure}
In our implementation, we construct separate graph for each country’s industrial sectors, and these graphs share the same architecture and parameters, treating country as a batch dimension. Each sector-node is connected to others based on domain knowledge and data-driven criteria~\cite{antonopoulos2020artificial}: for example, power generation node connected to all other industry nodes, since power supply affects widely and sectors historically exhibit correlated emission trends. We also estimate self-loops on each node in order to preserve node-specific information. Edges can be weighted. We initialize edge weights using Pearson correlation of past emission time series between sectors, which first attempt to capture historical interdependence. We also test with unweighted edges defined by threshold (e.g. connect if correlation > 0.5). The adjacency matrix $A(t)$ therefore represents a dynamic graph, for simplicity, we start with static graph capturing average relationships over the dataset period.

\subsection{Graph Neural Network with Attention}
\begin{figure}[t]
  \centering
  \includegraphics[scale=0.25]{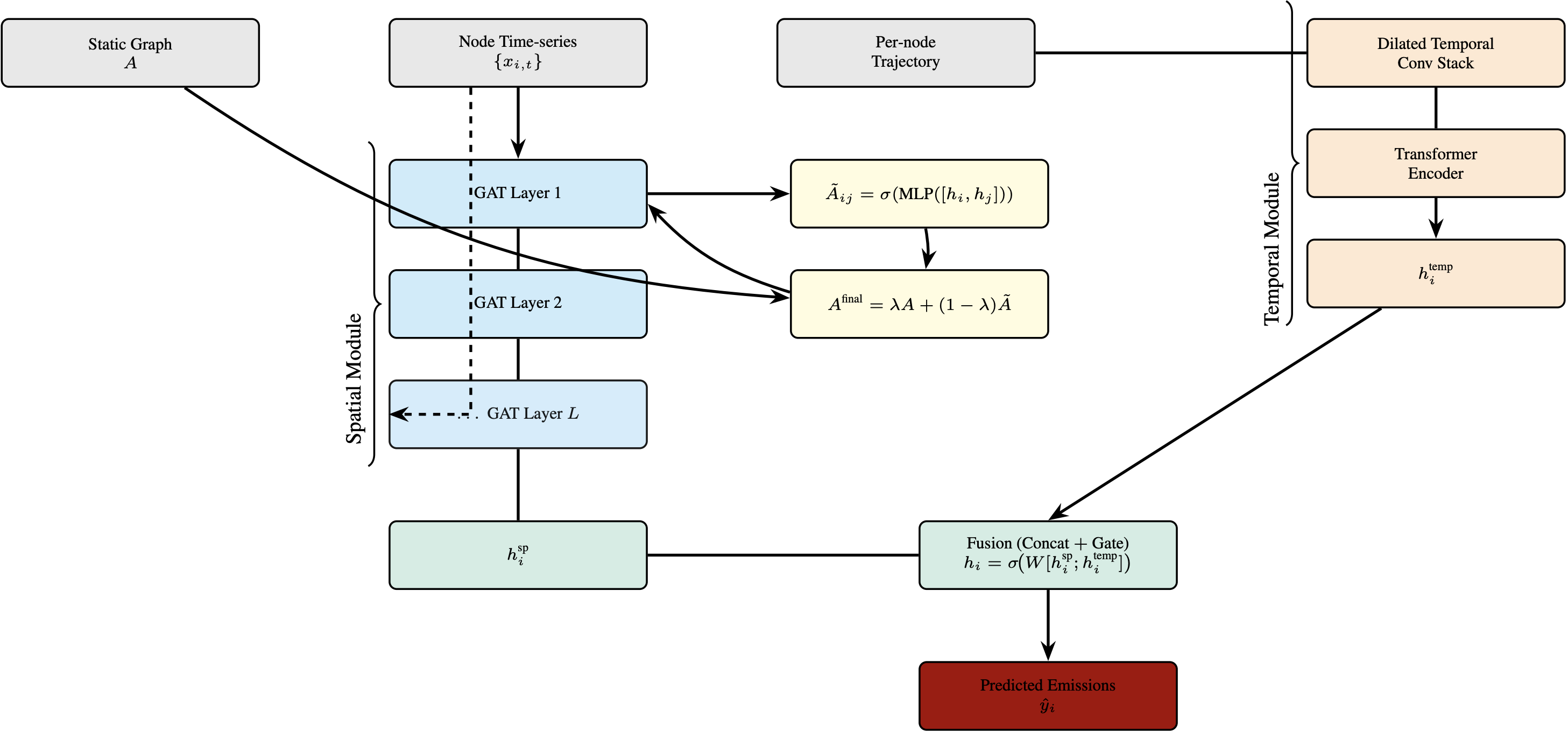}
  \caption{DGL Spatio‑temporal architecture combining multi‑layer Graph Attention Network with multi‑scale Temporal Transformer to predict node‑level industrial emissions.}
\label{fig:wide_gat_transformer}
\end{figure}

Given the constructed graph, we employ Graph Attention Network (GAT) as the spatial module. GAT layer allows node $i$ computes attention coefficients $\alpha_{ij}$ for each neighbor $j \in \mathcal{N}(i)$ based on similarity of node features from node embedding. These weights are normalized across neighbors and used to compute a weighted sum of neighbor messages. It lets the model learn which industries most strongly influence node $i$’s emissions. For instance, “aluminum smelting” attends strongly to the “power generation” since electricity is major input, whereas the “textile” attends more to “chemicals” and “transportation” which share dependencies. We stack $L$ graph attention layers to propagate information across multiple hops. In practice, we found $L=2$ or $3$ sufficient to capture indirect effects without over-smoothing. We also include skip connections to preserve original features. The initial feature of each node fed into GNN is derived from the node’s time-series (e.g. the latest year’s data and summary statistics of its feature history). In parallel, the temporal module will be processing the sequence, which we integrate them later, see Figure \ref{fig:wide_gat_transformer}.

One innovation in our spatial stage is using dynamic adjacency matrix learned and updated during training. Instead of fixed graph, we allow the model to adjust edge weights. We implement this by computing adjacency: $\tilde{A}_{ij} = \sigma( \text{MLP}([h_i, h_j]) )$ where $h_i, h_j$ are node embeddings from the previous layer and $\sigma$ is sigmoid kernel. This learned $\tilde{A}$ captures evolving relationships – for example, if at a certain time two industries become more coupled due to policy or market changes, the model can increase their connection weight. We use weighted combination of the static prior $A$ and the learned $\tilde{A}$ to stabilize training.
\subsection{Temporal Sequence Modeling}
Each node has a sequence of feature vectors over time: $X_i = {x_{i,t_1}, x_{i,t_2}, ..., x_{i,t_T}}$. These features include energy consumption by type, production output, economic indicators, etc., and the target emissions for supervised training. To model this sequence, we incorporate Temporal Module. Our chosen architecture uses multi-scale Temporal Convolution + Transformer, extracting short-term patterns with dilated convolution module, then with transformer encoder on top to capture long-term trends. The input is the sequence of single node’s feature trajectory, with positional encodings for the time dimension (e.g. year indices). The transformer outputs contextualized embeddings for each node at time. 

Because vanilla transformer would treat nodes independently focusing on time, we combine it with GNN outputs. Specifically, we feed per-node time series into the transformer to get temporal embedding $h^{(temp)}_i$, and combine this with the spatial embedding $h^{(sp)}_i$ from GNN. The fusion is done via concatenation followed by fusion layer with gating mechanism that weights spatial vs temporal information.

The output of the temporal module is a representation of each node’s time series which then informs the prediction.

\subsection{Prediction Layer}
After obtaining embeddings that incorporate both spatial and temporal context, we feed these into final prediction head. We train the model in a supervised fashion with Mean Squared Error (MSE) loss between predicted and actual emissions. We also experimented with a multi-task setup: predicting not just total emissions but also intermediate quantities (e.g. emissions from specific fuel types) to encourage the model to learn a richer representation, but ultimately focus on the primary target of total $CO_2$. During training, we activate appropriate mask for transformers to avoid using future data. We split data into training/validation/test by time and by country to test generalization.

\subsection{Alignment with Policy Goals}

We incorporate policy context in multiple ways. The graph design first reflect policy structure: e.g. connecting industries within a country emphasizes national policies link, whereas correlated same industries across countries can help identify if a certain sector globally is lagging or leading. Additionally, we frame part of the learning as multi-objective problem: not just minimizing prediction error, but also maximizing interpretability and policy relevance. Practically, we evaluate interpretability by measuring sparsity of attention, and by soliciting expert judgment on whether the explanations make sense. 

\section{Experiment}
\subsection{Dataset}
To ensure the study’s global relevance and reproducibility, we use publicly available data on industrial carbon emissions. The primary dataset comes from the Emissions Database for Global Atmospheric Research (EDGAR), version 8.0, released by the European Commission’s Joint Research Centre~\cite{crippa2024insights}. EDGAR provides estimates of $CO_2$ and other GHG emissions for every country, disaggregated by sector and fuel type, annually. Specifically, we extract $CO_2$ emissions from fossil fuel combustion and industrial processes for a range of sectors. From EDGAR’s sectoral breakdown, we focus on sectors that correspond to industries, including Electricity and Heat Production, Manufacturing (subdivided into Iron \& Steel, Chemicals, Cement, etc.), Transportation, Buildings/Construction, and Other Industry. We consider data for the years 1990 through 2020 (30 years), which covers the period of significant emission growth in emerging economies. The dataset spans 100+ countries. For manageability in modeling, we select a representative subset of countries across different development levels and emission profiles: e.g. China, United States, India, Germany, Brazil, South Africa, etc. In addition to emissions, we compile features for each country-sector-year entry by Energy consumption by fuel type, Industrial output indicators, and Policy or technology indicators. The combined dataset thus forms a panel: each record identified by (country, sector, year) with a set of input features and an output ($CO_2$ emissions for that (country, sector) in that year).

After preprocessing, the training dataset includes N = 10 countries × 50 sectors × 26 years = 13000 samples, and similarly sized validation and test sets. While the absolute size is modest, the richness comes from each sample having $\approx$10-20 features and our graph connecting these samples across sectors. It’s worth noting that the dataset covers a diverse range: e.g., China’s industrial emissions tripled over 1990-2015, whereas the EU’s slightly declined – giving the model instances of both growth and mitigation. Such diversity helps the model avoid overfitting to one trend. The EDGAR data has been used in other high-profile analyses, lending credibility and comparability to our results.

Finally, to validate our data pipeline, we compare aggregated model inputs to known totals. For example, summing the sector emissions we use for a country yields that country’s total emissions, which we cross-checked with World Bank figures. This gives confidence that our data represents reality well, and any insights derived could be meaningful for actual policy.

\subsection{Experimental Setup}
We implemented our model using PyTorch Geometric for the GNN components and the standard PyTorch library for Transformers. Training was performed on an NVIDIA A100 GPU, converged within 50 epochs (around 2 hours for the largest configuration). We used the Adam optimizer with an initial learning rate of 0.001. Early stopping was employed based on validation $R^2$ to prevent overfitting.

We compare our proposed DGL model against several baselines: \ding{182} Multiple Linear Regression (MLR): A simple linear regression on all regularization features. This tests if linear models with all features can handle multicollinearity and provides interpretability. \ding{183} Elastic Net Regression: Similar to MLR but with L1 and L2 penalties tuned via cross-validation. \ding{184} Cluster Regression (DPR)~\cite{zhang2025multicollinearity}: The two-step baseline where we first cluster data points and then fit separate elastic net models per cluster. We use DBSCAN to cluster the training samples based on their features; the number of clusters found was around 8 for our data. This approach explicitly addresses multicollinearity by separation. \ding{185} Multilayer Perceptron (MLP): Feed-forward neural network that takes concatenation of all features and predicts emissions. This ignores graph or sequence structure, effectively treating each sample independently with a universal approximator. We used 3 hidden layers with ReLU activations. \ding{186} LSTM (seq-only): Many-to-one LSTM model per entity that produces prediction for the final time step. This accounts for time but not interactions between entities. \ding{187} Transformer (seq-only): Transformer encoder that attends across time for each entity without cross-entity attention. \ding{188} GNN-static: Graph Neural Network that uses the graph and static features without temporal component. Essentially, this predicts emissions from a single snapshot using neighbor information, testing if graph alone offers improvement.

For each model, we perform hyperparameter tuning. For deep models, we tune hidden sizes (64 as base), number of layers (2 graph conv layers, 2 LSTM layers or 2 transformer layers), and dropout (0.1–0.3) to balance bias and variance. Regression were tuned for their regularization parameter $\lambda$ and cluster $\epsilon$ for DBSCAN. See detail in Appendix.

We primarily evaluate with Coefficient of Determination ($R^2$) and Root Mean Squared Error (RMSE) on the test set, with Mean Absolute Percentage Error (MAPE) provides an intuitive error magnitude. We also measure computational performance (training time, inference time) since a model that is too slow would hinder interactive policy use. DGL model, being more complex, is slower than simple regression but still provides results in seconds for a given scenario, which is acceptable.

\subsection{Overall Prediction Performance}
\begin{table}[t!]
  \centering
  \caption{Model Performance Comparison on Industrial $CO_2$ Emissions (Test Set).}
  \label{tab:my_label3}
  \setlength{\tabcolsep}{4.5mm}
  \resizebox{0.5\textwidth}{!}{
  \begin{tabular}{lccc}
    \toprule
    Model & $R^2$ & RMSE ($MtCO_2$) & MAPE (\%) \\
    \midrule
    Multiple Linear Reg. & 0.72 & 220 & 18.4 \\
    Elastic net Reg. & 0.85 & 150 & 12.7 \\
    DPR & 0.88 & 130 & 10.5 \\
    MLP & 0.80 & 180 & 15.0  \\
    LSTM & 0.90 & 120 & 9.8  \\
    Transformer & 0.91 & 115 & 9.5  \\
    GNN & 0.86 & 140 & 11.3  \\
    \textbf{DGL} & \textbf{0.95} & \textbf{90} & \textbf{7.8}\\
    \bottomrule
  \end{tabular}
  }
\end{table}
Table \ref{tab:my_label3} summarizes the performance of our model versus baselines on the test set (2019–2024 data for selected countries/sectors). Our Deep Graph Learning model achieved the highest accuracy in terms of $R^2$ (0.95) and lowest error in RMSE (90) and MAPE, meaning it explains 95\% of the variance in the industrial emissions data – a significant improvement over traditional architecture. In practical terms, RMSE of DGL is roughly 25\% lower than LSTM or Transformer alone. It’s worth noting MLP underperformed Elastic Net; likely the limited data and multicollinearity hampered the simple neural network, causing it to overfit or get confused by correlated inputs.

Notably, our model’s performance exceeds state-of-the-art research findings. For instance, LSTM-GAT for carbon emissions achieved 89.5\% accuracy~\cite{wu2024carbon}. Another transformer-based study on regional emissions improved errors by 4-8\% over baselines~\cite{liu2023multi} – our results show a larger improvement, possibly because industrial data is highly structured and benefited more from graph modeling. We provide robustness tests in Appendix.

We also evaluated generalization: our model trained on a set of countries predicted emissions for unseen countries with reasonable accuracy (R² 0.9). This indicates the model learned transferable representation about how industries behave, rather than just memorizing each country’s trend. In contrast, traditional models struggle with such transfer without retraining. We also show ablation studies in Appendix.

\subsection{Learned Graph Patterns}
\begin{figure}[t]
  \centering
  \includegraphics[scale=0.4]{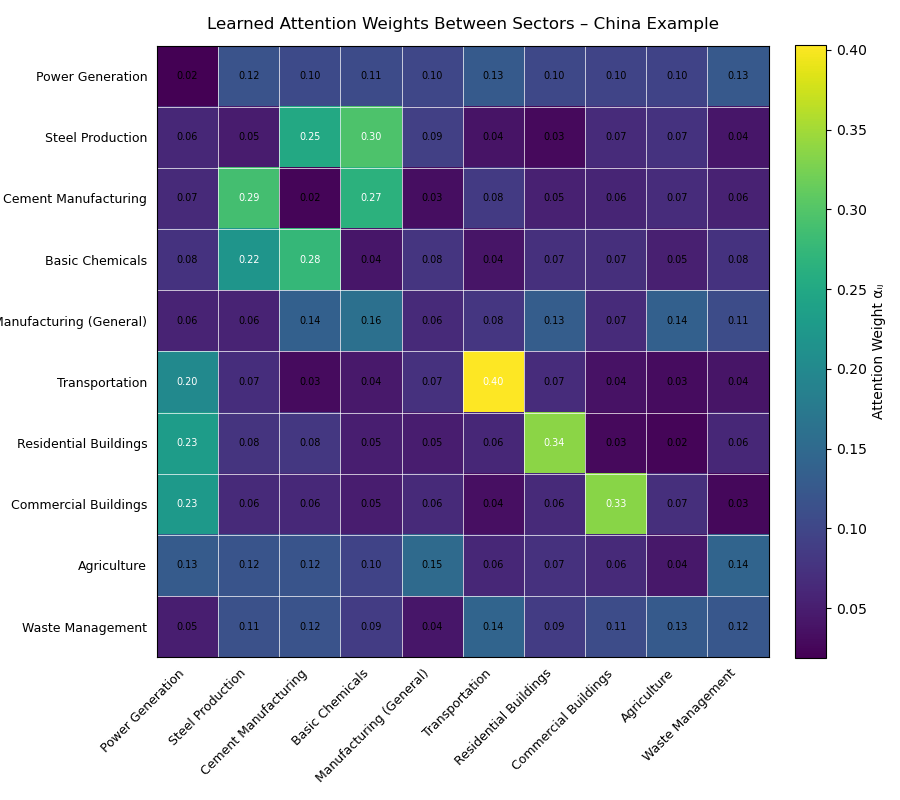}
  \caption{\textbf{Attention weight matrix ($\alpha_{ij}$) learned for China.}}
\label{figure4}
\end{figure}

One insightful result is the structure the model learned in the graph. Figure \ref{figure4} visualizes the attention weight matrix $\alpha_{ij}$ between sectors for one example country (China) as learned by the final model. Several intuitive patterns emerge:

\ding{182} The Power generation sector has high outgoing influence to almost all other sectors. The attention weights from Power to industries like Steel, Chemicals, Manufacturing are among the largest. This aligns with reality: access to electricity (often coal-based in China’s timeframe) is a limiting factor for those industries’ emissions. Essentially, the model learned that predicting emissions in manufacturing requires knowing how much power sector is emitting (since more power generation implies more industrial electricity use and thus indirectly emissions in that sector’s supply chain). This demonstrates the model’s ability to capture a kind of scope 2 emissions effect.

\ding{183} There is a cluster of heavy industries (Steel, Cement, Chemicals) that attend strongly to each other. These sectors share partial common drivers (e.g. they all use coal and are energy-intensive). The model likely leveraged that, attributing to broad economic growth or energy price conditions.

\ding{184} Transport and buildings sectors were somewhat isolated nodes – their strongest attention weights were mainly self-directed since transport rely on oil which doesn’t directly connect to other industries, and buildings emissions are mostly from power/heat consumption. It effectively left reasonable decoupled, reflecting the true structure.

\ding{185} The learned edges also had temporal dynamics: for instance, the influence of Power -> Industry was even higher in later years than earlier, reflecting increased electrification of industry in recent years. Our dynamic graph allowed such change, which a static one would miss. A detailed temporal pattern analysis can be found in Appendix.

These patterns give us confidence that the GNN component is learning meaningful relationships, not arbitrary ones. It’s notable that many of these relationships correspond to what one would find in an expert-crafted sectoral interaction matrix, but here it was derived from data and the prediction task.

\subsection{Case Study: Sectoral Intervention Simulations}
Beyond raw accuracy, we present a case study to demonstrate the model’s use for policy. We took the trained model for India’s industries and simulated a scenario: What if India’s coal consumption in heavy industries is reduced by 20\% starting in 2022, replaced by cleaner energy? Using our causal analysis module, we input a modified feature sequence for 2022-2025 where “Coal” for sectors like steel and cement was 20\% lower (and “Electricity” slightly higher to assume substitution). The model predicted India’s industrial emissions in 2025 would be 8\% lower than the baseline prediction with this intervention. The reduction in specific sectors was larger (15\% in cement, 12\% in steel), but was offset by slight increases elsewhere or economy-wide feedback captured by the model (e.g. power sector emissions rose due to more electricity usage, though cleaner per unit). This kind of simulation illustrates how DGL model can help quantify the impact of targeted policies on future emissions.

Another scenario: What if an economic slowdown reduces output in all sectors by 10\%? The model projected an 8\% emissions reduction, suggesting non-linear effects and fixed emissions portions. Such insights can guide expectations for economic vs. emission decoupling.

\section{Discussion}
One of the primary aims of this research is to ensure the AI system is interpretable and yields insights that can inform policy decisions for industrial decarbonization. We delve here into what the model has taught us about emission drivers and how these findings could shape interventions. We also discuss the model’s role in a policy-making context, addressing issues of fairness, usability, and limitations.

\subsection{Interpretability of the Model’s Insights}
\begin{figure}[t]
  \centering
  \includegraphics[scale=0.45]{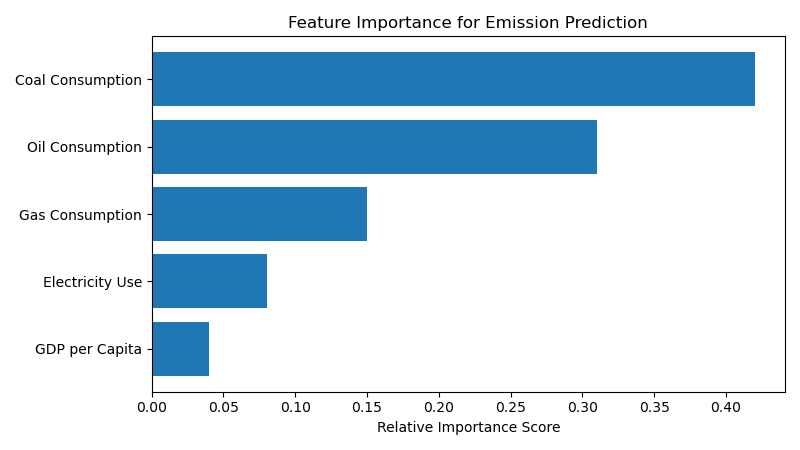}
  \caption{\textbf{Feature importance for emission prediction.} The relative importance scores of key features as determined by explainability analysis. “Coal Consumption” and “Oil Consumption” emerge as the most influential factors, which aligns with domain expectations that coal and oil are carbon-intensive fuels. “Gas Consumption” also has a moderate impact, while “Electricity Use” and a macro-economic indicator “GDP per Capita” show contributions. Such insights allow policymakers to identify which factors (fuel types, economic drivers) are most critical to address for reducing emissions.}
\label{figure5}
\end{figure}

The combination of attention mechanisms and post-hoc analysis provides a rich picture of the factors influencing emissions:

\ding{182} Feature Importance Recap: As shown in Figure \ref{figure5}, the model identified fossil fuel use as the dominant contributor to industrial $CO_2$ emissions, which is unsurprising but quantitatively confirms the narrative. It also gave importance to interactions that, in some heavy industries, it wasn’t just “coal amount” but the ratio of coal to output (emissions intensity) that determined emissions.

\ding{183} Inter-Industry Links: The graph attention revealed clusters of industries whose emissions are intertwined. For policymakers, this suggests that a systems approach is needed – regulating one industry will have ripple effects. For example, improving efficiency in the power sector (power generation emissions factor) will propagate to reduce predicted emissions in steel, manufacturing, proportionally to their attention weights to power, benefiting all connected industries. Our model effectively highlights these leverage points. Thus, the attention weights can be interpreted as elasticities or sensitivities. This information could be valuable in integrated assessment models or cross-sector collaboration initiatives.

\ding{184} Temporal Dynamics: Through the transformer, the model can indicate when an industry’s emissions trajectory changed significantly. Recognizing such inflection points is important for policy timing: it suggests when and possibly why an industry decoupled emissions from growth. Policymakers can investigate those periods (e.g., was there a policy that caused that? or an external shock?) and either replicate success or avoid pitfalls from other periods.

\subsection{Policy Implications}
From the model’s outcomes, several policy implications can be drawn:

\ding{182} Targeting Key Sectors and Fuels: The model clearly points to certain sector-fuel combinations as “low-hanging fruit” for emissions reduction. Coal use in industrial sectors stands out. This reinforces existing calls to phase out coal, but our results can prioritize which sectors to tackle first. Steel and power are most coal-dependent and influential across the graph, focusing on those (through measures like coal-to-gas fuel switching, CCS technology for steel) will yield the largest immediate benefits. On the other hand, sectors like textiles appear lower priority for heavy mitigation investment that its emissions will fall if upstream power cleans up.

\ding{183} Holistic Strategies – Avoid Policy Siloes: Our model implies that isolated policies might not suffice. For example, an industrial energy efficiency program will not achieve full potential if the power sector remains dirty; similarly, cleaning the power grid but ignoring industry efficiency will leave potential gains on the table. A coordinated policy that addresses power generation and industrial usage simultaneously would maximize impact, a concept supported by the connected nature of our model’s graph. This argues for integrated energy and industrial policies.

\ding{184} Fairness and Equity: Our approach allows the examination of fairness in emissions allocation. For instance, we can assess proposals like “each industry should reduce emissions by X\%” or “each country gets a certain emissions quota for industry.” and evaluate outcomes: which sectors or countries struggle more in our predictions under those constraints? Our analysis suggests that uniform cuts are not equally realistic: sectors differ in abatement capacity. Heavy industries have less easy alternatives, whereas others can electrify or improve efficiency more. Thus fairness might mean heavy industries get a slightly higher allowed emissions for longer, while easier-to-abate sectors take more early action. Also, across countries, our results show developing countries’ industrial emissions are rising faster mainly due to growth, even if intensity improves. This supports the principle of common but differentiated responsibilities – developed nations should help by funding cleaner tech transfers so developing nations can grow without high emissions. We can simulate such assistance: e.g., lowering the emissions intensity feature for a developing country’s steel sector to match developed country levels and seeing the emissions difference.

\ding{185} Monitoring and Verification: The model could be used by policymakers as a monitoring tool. Because it links factors to emissions, if a country implements a policy, the model can be used with updated inputs (post-policy energy use patterns) to verify if emissions are trending as expected. If not, compliance is an issue or rebound effects occurred. Essentially, it can serve as a digital twin for industrial emissions: a mirror of the real system that can be poked and prodded to see responses.

\ding{186} Sustainable Development Goals (SDGs): Our work has implications for multiple SDGs: SDG 9 (Industry, Innovation, and Infrastructure) by promoting innovation in industrial processes; SDG 7 (Affordable and Clean Energy) by highlighting the need for cleaner fuel mix; and SDG 13 (Climate Action). One particular angle is just transition – ensuring workers and communities in high-emission industries are not left behind. While our model doesn’t directly address socioeconomic outcomes, by identifying which industries will likely need to decline (e.g., coal-heavy ones), it flags where transition plans (job retraining, economic diversification) are needed. Policymakers can use this to proactively invest in those areas.

We also state Fairness Considerations, Limitations, and Broader Social Impact in Appendix.
\section{Conclusion}
We presented a novel graph-based deep learning framework for analyzing and predicting industrial carbon emissions, aiming to support climate action in the industrial sector. Our approach addresses key technical challenges – multicollinearity among features and complex interdependencies across sectors and time by integrating Graph Neural Networks, temporal transformers, and causal inference tools. More importantly, it provides interpretable insights: identifying which fuels and industries are driving emissions, how different sectors influence each other, and what the likely outcomes of certain interventions would be. We emphasize that achieving social impact requires not only accuracy but also trustworthiness of AI – hence our focus on explainability and policy alignment. The framework can be extended to other sectors or environmental indicators, suggesting a broad applicability of deep graph learning in sustainability domains.

These insights translate into concrete policy guidance, reinforcing that phasing out coal from industry, electrifying processes, and cleaning the power grid are high-impact steps. We also highlight the importance of coordinated policies across sectors. We showed how the model can inform fair allocation of emissions reductions, aligning with global climate justice discourse, by quantifying different capacities and responsibilities of sectors and regions. This ensures that AI recommendations can be used in equitable climate strategies, addressing SDG13 and beyond.
\bibliography{custom}

@article{ritchie2020breakdown,
  title={Breakdown of carbon dioxide, methane and nitrous oxide emissions by sector},
  author={Ritchie, Hannah and Rosado, Pablo and Roser, Max},
  journal={Our World in Data},
  year={2020}
}

@article{change2022mitigating,
  title={Mitigating Climate Change},
  author={Change, Climate},
  journal={Working Group III contribution to the sixth assessment report of the intergovernmental panel on climate change},
  year={2022},
  publisher={Springer}
}

@article{zhang2025multicollinearity,
  title={Multicollinearity Resolution Based on Machine Learning: A Case Study of Carbon Emissions},
  author={Zhang, Xuanming},
  journal={arXiv preprint arXiv:2507.02912},
  year={2025}
}

@article{ahmed2023deep,
  title={Deep learning modelling techniques: current progress, applications, advantages, and challenges},
  author={Ahmed, Shams Forruque and Alam, Md Sakib Bin and Hassan, Maruf and Rozbu, Mahtabin Rodela and Ishtiak, Taoseef and Rafa, Nazifa and Mofijur, M and Shawkat Ali, ABM and Gandomi, Amir H},
  journal={Artificial Intelligence Review},
  volume={56},
  number={11},
  pages={13521--13617},
  year={2023},
  publisher={Springer}
}

@article{wang2023forecasting,
  title={Forecasting power demand in China with a CNN-LSTM model including multimodal information},
  author={Wang, Delu and Gan, Jun and Mao, Jinqi and Chen, Fan and Yu, Lan},
  journal={Energy},
  volume={263},
  pages={126012},
  year={2023},
  publisher={Elsevier}
}

@article{wen2023modeling,
  title={Modeling and forecasting CO2 emissions in China and its regions using a novel ARIMA-LSTM model},
  author={Wen, Tingxin and Liu, Yazhou and he Bai, Yun and Liu, Haoyuan},
  journal={Heliyon},
  volume={9},
  number={11},
  year={2023},
  publisher={Elsevier}
}

@article{cowls2023ai,
  title={The AI gambit: leveraging artificial intelligence to combat climate change—opportunities, challenges, and recommendations},
  author={Cowls, Josh and Tsamados, Andreas and Taddeo, Mariarosaria and Floridi, Luciano},
  journal={Ai \& Society},
  volume={38},
  number={1},
  pages={283--307},
  year={2023},
  publisher={Springer}
}

@article{york2003stirpat,
  title={STIRPAT, IPAT and ImPACT: analytic tools for unpacking the driving forces of environmental impacts},
  author={York, Richard and Rosa, Eugene A and Dietz, Thomas},
  journal={Ecological economics},
  volume={46},
  number={3},
  pages={351--365},
  year={2003},
  publisher={Elsevier}
}

@article{tao2024dynamics,
  title={Dynamics between energy intensity and carbon emissions: What does the clustering effect of labor and capital play?},
  author={Tao, Miaomiao and Wen, Le and Sheng, Mingyue Selena and Yan, Zheng Joseph and Poletti, Stephen},
  journal={Journal of Cleaner Production},
  volume={452},
  pages={142223},
  year={2024},
  publisher={Elsevier}
}

@article{kim2019multicollinearity,
  title={Multicollinearity and misleading statistical results},
  author={Kim, Jong Hae},
  journal={Korean journal of anesthesiology},
  volume={72},
  number={6},
  pages={558--569},
  year={2019},
  publisher={Korean Society of Anesthesiologists}
}

@article{chan2022mitigating,
  title={Mitigating the multicollinearity problem and its machine learning approach: a review},
  author={Chan, Jireh Yi-Le and Leow, Steven Mun Hong and Bea, Khean Thye and Cheng, Wai Khuen and Phoong, Seuk Wai and Hong, Zeng-Wei and Chen, Yen-Lin},
  journal={Mathematics},
  volume={10},
  number={8},
  pages={1283},
  year={2022},
  publisher={MDPI}
}

@article{zhang2023multicollinearity,
  title={Multicollinearity Resolution Based on Machine Learning: A Case Study of Carbon Emissions in Sichuan Province},
  author={Zhang, Xuanming and Wang, Xiaoxue and Chen, Yonghang},
  journal={arXiv preprint arXiv:2309.01115},
  year={2023}
}

@article{gao2023review,
  title={A review of building carbon emission accounting and prediction models},
  author={Gao, Huan and Wang, Xinke and Wu, Kang and Zheng, Yarong and Wang, Qize and Shi, Wei and He, Meng},
  journal={Buildings},
  volume={13},
  number={7},
  pages={1617},
  year={2023},
  publisher={MDPI}
}

@article{waqas2024critical,
  title={A critical review of RNN and LSTM variants in hydrological time series predictions},
  author={Waqas, Muhammad and Humphries, Usa Wannasingha},
  journal={MethodsX},
  volume={13},
  pages={102946},
  year={2024},
  publisher={Elsevier}
}

@article{agga2022cnn,
  title={CNN-LSTM: An efficient hybrid deep learning architecture for predicting short-term photovoltaic power production},
  author={Agga, Ali and Abbou, Ahmed and Labbadi, Moussa and El Houm, Yassine and Ali, Imane Hammou Ou},
  journal={Electric Power Systems Research},
  volume={208},
  pages={107908},
  year={2022},
  publisher={Elsevier}
}

@article{mumuni2025automated,
  title={Automated data processing and feature engineering for deep learning and big data applications: a survey},
  author={Mumuni, Alhassan and Mumuni, Fuseini},
  journal={Journal of Information and Intelligence},
  volume={3},
  number={2},
  pages={113--153},
  year={2025},
  publisher={Elsevier}
}

@article{wu2024carbon,
  title={Carbon emissions forecasting based on temporal graph transformer-based attentional neural network},
  author={Wu, Xingping and Yuan, Qiheng and Zhou, Chunlei and Chen, Xiang and Xuan, Donghai and Song, Jinwei},
  journal={Journal of Computational Methods in Sciences and Engineering},
  volume={24},
  number={3},
  pages={1405--1421},
  year={2024},
  publisher={SAGE Publications Sage UK: London, England}
}

@article{wang2024multiscale,
  title={Multiscale graph based spatio-temporal graph convolutional network for energy consumption prediction of natural gas transmission process},
  author={Wang, Chen and Zhou, Dengji and Wang, Xiaoguo and Liu, Song and Shao, Tiemin and Shui, Chongyuan and Yan, Jun},
  journal={Energy},
  volume={307},
  pages={132489},
  year={2024},
  publisher={Elsevier}
}

@article{antonopoulos2020artificial,
  title={Artificial intelligence and machine learning approaches to energy demand-side response: A systematic review},
  author={Antonopoulos, Ioannis and Robu, Valentin and Couraud, Benoit and Kirli, Desen and Norbu, Sonam and Kiprakis, Aristides and Flynn, David and Elizondo-Gonzalez, Sergio and Wattam, Steve},
  journal={Renewable and Sustainable Energy Reviews},
  volume={130},
  pages={109899},
  year={2020},
  publisher={Elsevier}
}

@article{crippa2024insights,
  title={Insights into the spatial distribution of global, national, and subnational greenhouse gas emissions in the Emissions Database for Global Atmospheric Research (EDGAR v8. 0)},
  author={Crippa, Monica and Guizzardi, Diego and Pagani, Federico and Schiavina, Marcello and Melchiorri, Michele and Pisoni, Enrico and Graziosi, Francesco and Muntean, Marilena and Maes, Joachim and Dijkstra, Lewis and others},
  journal={Earth System Science Data},
  volume={16},
  number={6},
  pages={2811--2830},
  year={2024},
  publisher={Copernicus GmbH}
}

@inproceedings{liu2023multi,
  title={Multi-Scale Graph Transformer Networks for Regional Emission Spatiotemporal Prediction},
  author={Liu, Qianming and Pei, Lihong and Cao, Yang},
  booktitle={2023 9th International Conference on Big Data and Information Analytics (BigDIA)},
  pages={436--441},
  year={2023},
  organization={IEEE}
}

@article{williges2022fairness,
  title={Fairness critically conditions the carbon budget allocation across countries},
  author={Williges, Keith and Meyer, Lukas H and Steininger, Karl W and Kirchengast, Gottfried},
  journal={Global Environmental Change},
  volume={74},
  pages={102481},
  year={2022},
  publisher={Elsevier}
}

@article{zhang2024seeker,
  title={Seeker: Enhancing Exception Handling in Code with LLM-based Multi-Agent Approach},
  author={Zhang, Xuanming and Chen, Yuxuan and Yuan, Yuan and Huang, Minlie},
  journal={arXiv preprint arXiv:2410.06949},
  year={2024}
}

@article{li2024evocodebench,
  title={Evocodebench: An evolving code generation benchmark aligned with real-world code repositories},
  author={Li, Jia and Li, Ge and Zhang, Xuanming and Dong, Yihong and Jin, Zhi},
  journal={arXiv preprint arXiv:2404.00599},
  year={2024}
}

@article{wang2024theoretical,
  title={Theoretical Proof that Auto-regressive Language Models Collapse when Real-world Data is a Finite Set},
  author={Wang, Lecheng and Shi, Xianjie and Li, Ge and Li, Jia and Zhang, Xuanming and Dong, Yihong and Jiao, Wenpin and Mei, Hong},
  journal={arXiv preprint arXiv:2412.14872},
  year={2024}
}

@article{zhang2025metamind,
  title={MetaMind: Modeling Human Social Thoughts with Metacognitive Multi-Agent Systems},
  author={Zhang, Xuanming and Chen, Yuxuan and Yeh, Min-Hsuan and Li, Yixuan},
  journal={arXiv preprint arXiv:2505.18943},
  year={2025}
}

@article{sabour2025human,
  title={Human decision-making is susceptible to ai-driven manipulation},
  author={Sabour, Sahand and Liu, June M and Liu, Siyang and Yao, Chris Z and Cui, Shiyao and Zhang, Xuanming and Zhang, Wen and Cao, Yaru and Bhat, Advait and Guan, Jian and others},
  journal={arXiv preprint arXiv:2502.07663},
  year={2025}
}

@misc{donggeneralization,
    title   = {Generalization or Memorization: Evaluating Data Contamination for Large Language Models},
    author  = {Dong, Yihong and Jiang, Xue and Zhang, Xuanming and Liu, Huanyu and Jin, Zhi and Gu, Bin and Yang, Mengfei and Li, Ge},
    journal = {Stanford Computer Science},
    year    = {2025},
    month   = {Jan},
    url     = {https://web.stanford.edu/~zhangxm/Generalization_or_Memorization__Evaluating_Data_Contamination_for_Large_Language_Models.pdf},
    note    = {\url{https://web.stanford.edu/~zhangxm/Generalization_or_Memorization__Evaluating_Data_Contamination_for_Large_Language_Models.pdf}}
}
\bibliographystyle{unsrt}

\newpage
\appendix

\section*{Appendix}

\section{Training Procedure and Hyperparameters}
\label{app1}
Key Hyperparameters:

GNN: 2 layers GAT with 4 attention heads each, embedding size 64 per head (concatenated to 256 per layer output), LeakyReLU activation, dropout 0.2 on edges.

LSTM: 2 layers, hidden size 128, sequence length = 10 years.

Transformer: 4 heads, 2 layers, hidden size 128, position encoding for up to 30 years.

Fusion: We concatenated GNN output and LSTM output (each 128-d) to 256-d then passed through a 64-d dense layer.

Loss: MSE loss on log emissions. For some runs we added L1 regularization on attention weights to encourage sparsity (to enhance interpretability by having fewer strong connections), with a small coefficient $10^{-5}$.

We ensured all models saw the same training data and same train/test split for fairness. For each baseline, we report the best performance achieved after tuning. Each experiment is run 5 times with different random seeds to ensure robustness; we report average metrics and note the variance if significant.

\section{Ablation Studies}
We conducted ablation experiments to understand the contribution of each component in DGL:

\ding{182} Removing the graph (i.e., no inter-industry edges, each sector predicted independently) dropped R² by about 4 points (from 0.95 to 0.91). This demonstrates the graph contributes significantly to predictive power by providing additional context. It also increased error specifically in cases where industries are known to influence each other (e.g., power and others), confirming that relationships were being exploited.

\ding{183} Removing the temporal module (only using static features and graph) had an even larger effect, dropping R² to ~0.85. This is expected since temporal trends are strong; it shows both aspects (spatial and temporal) are essential, with temporal slightly more so for pure prediction.

\ding{184} Using a static, non-learnable adjacency vs. our dynamic adjacency led to a drop in accuracy (0.93 R² vs 0.95). The model with dynamic adjacency adapted better, especially in later years when partial relationships changed. For instance, it learned that some industries became more electrified over time, effectively strengthening the edge between those industries and the power sector node as time progressed, which a static graph could not do.

\ding{185} If we disable the interpretability regularization (attention L1), the accuracy was essentially unchanged, indicating those constraints did not harm performance notably. However, the resulting attention patterns were denser, making explanations harder. With regularization, each industry node tended to have 1-3 strong connections instead of being moderately connected to many, which is preferable for interpretation.

\section{Temporal Pattern Analysis}
The transformer module yielded interesting temporal attention insights. We found that for most sectors, the model pays most attention to the most recent 3-5 years of data when predicting the next year, which makes sense since recent trends matter. However, in certain cases, it also gave notable weight to years where significant changes occurred. For example, in U.S. industrial emissions, the year 2009 (a recession year) had an outsize attention weight for predicting 2010 emissions, likely because the sharp drop and rebound around that period is a critical pattern to model. Likewise, for China, 2001-2002 got higher attention in some sectors (entry to WTO, rapid industrial growth phase), indicating the model recognized a regime shift in that period.

We also examined seasonality and periodicity, specifictly, looking for policy cycles (e.g. five-year plans in China). There was evidence the DGL model picked up the right five-year cycle: e.g., attention spikes at those intervals for certain features aligning with policy implementation cadence, though this is an area for further exploration.

\section{Robustness Tests}
We tested robustness under two conditions: data noise and extreme events. For data noise, we added random Gaussian noise to 10\% of input features during model testing to see if predictions change drastically. DGL model proved robustness – R² dropped only 0.01, whereas simpler models like linear regression dropped 0.03. The graph structure likely helped smooth out noise (neighbor info compensating) and the regularization in the model prevented reacting to odd spikes. Under extreme events, we examined how the model handled 2020 (the year of COVID-19 disruptions). Interestingly, when we included 2020 in the test set, the model slightly under-predicted the emission drop since the actual drop in some sectors was even larger due to lockdowns. This is understandable, as 2020 was outside the range of normal behavior. However, the model still got the direction and relative sectoral impacts right (e.g. it knew transport emissions would drop more than power emissions in 2020). This points to a limitation: truly unprecedented events (black swans) are hard for any model to predict without specific features indicating them. We try to discuss this and how model can be updated.

\section{Fairness Considerations}
We explicitly examine fairness in the context of model recommendations:

Our model can output, for each country, an approximate emissions reduction burden needed to meet a global target if we constrain with objective in scenario. We found that if each country’s industries improve at the historically best rates, the aggregate still overshoots Paris Agreement goals – meaning everyone needs to do more. But how to distribute the “more”? A naive approach would cut every country’s industrial emissions by, say, 20\%. Our model suggests this is feasible for some and very challenging for others, because frontier have already picked low-hanging fruits and others have not. Fairness dictates that countries which have not yet reaped easy gains should do so with support rather than forcing equal absolute cuts. This aligns with fairness principles discussed in climate policy literature~\cite{williges2022fairness}. Our model can quantify, for instance, how much more $CO_2$ per dollar of GDP one country emits than another in the same industry – a measure of efficiency and, indirectly, responsibility to improve. Those with high $CO_2/GDP$ in a sector arguably should reduce more with technologically aid. We can also flip it: those with lower $CO_2/GDP$ (efficient tech) can share technology to help others catch up, which our model would simulate as raising others’ efficiency feature, showing how much emissions could drop if such tech transfer occurred.

From an AI ethics perspective, we also consider fairness in the model itself. Did the model treat similar countries similarly, or was it biased by data volume? We noticed the model had slightly higher error on smaller countries. This could be a data imbalance issue caused by big emitters dominate training. Techniques like re-weighting were considered. For now, this highlights that one must be careful using the model for very small countries, trying aggregate them or accept larger uncertainty. We avoid any protected attributes, and state that regional biases (like focusing on developed vs. developing patterns) are features to monitor. The model is ultimately a tool; fairness comes in how its results are applied.

\section{Broader Social Impact}
On the positive side, our work demonstrates how advanced AI can directly contribute to addressing climate change – one of society’s grand challenges. It provides a template for combining deep learning with expert knowledge (graph structure, fairness considerations) to create tools that are powerful yet aligned with human values and policy frameworks. If adopted by analysts or governments, it could improve the evidence base for climate action, potentially leading to more effective and just policies, thereby reducing emissions faster and mitigating climate risks.

However, we must also consider potential misuse or misinterpretation. A sophisticated model might give a false sense of certainty. Policymakers could over-rely on it without understanding uncertainties, leading to policy mistakes. We advocate the model as a decision support tool, not a decision maker. It should be used in conjunction with stakeholder consultation. Additionally, ensuring transparency (open-sourcing the model and data, documenting assumptions) is vital for trust – something we have prioritized by using public data and describing the method in detail.

Lastly, the fairness aspect: any biases in data (e.g., underreported emissions, or lack of data from certain regions) could reflect in model outputs and inadvertently disadvantage or advantage some party. Recognizing and correcting for these biases is an ongoing effort, but at least with interpretable outputs, such issues are more likely to be spotted than in a black-box scenario.

\section{Limitations and Future Work}
While our framework shows strong performance and utility, there are limitations:

\ding{182} Data Quality and Granularity: Annual national-level data may mask sub-annual or sub-national variations. Industrial emissions often fluctuate monthly with production cycles. Incorporating higher frequency data or plant-level data could further improve model fidelity. Similarly, including more granular sector breakdowns could refine insights. The trade-off is complexity and data availability. In future work, we could attempt a hierarchical model~\cite{zhang2024seeker}: global -> national -> sector -> subsector, using appropriate graphs at each level.

\ding{183} Causality vs. Correlation: Our causal inference analysis is limited by being on the model, not the real world. If the model is wrong in partial causal assumption, the intervention results will be wrong. To improve this, we could integrate known causal structure (e.g., structural equation models alongside the network). Another approach is to use reinforcement learning or optimal control in tandem with the model to suggest optimal interventions, but ensuring the RL agent’s suggestions are fair and realistic would be challenging~\cite{li2024evocodebench}.

\ding{184} Dynamic Policy Environment: Policies themselves will change the relationships. If a major new regulation comes (e.g., a carbon tax), historical data won’t reflect its impact~\cite{wang2024theoretical,donggeneralization}. Our model can’t predict a novel policy impact without examples, unless we incorporate expert rules or simulation~\cite{zhang2025metamind}. One idea is to integrate economic model or climate-economic hybrid (IAM) to simulate policy impacts as additional training data. Alternatively, periodically retraining the model as new data (post-policy) comes in will be necessary. This is part of model maintenance in deployment.

\ding{185} Interpretability vs. Accuracy Trade-off: While we strived for both, there is always a balancing act. We added interpretability modules and regularizations that could be further enhanced, including directly optimizing for sparsity and rule extraction. Future work might explore extracting a simplified rule-based model from the deep model, as a form of distillation for policymakers who prefer rule lists. Ensuring these rules are accurate representations is tricky but would further build trust~\cite{sabour2025human}.

\ding{186} Scalability: Our experiments included up to 500 nodes (countries × sectors). In a scenario with all countries and many sectors, the graph could be much larger. Graph attention networks scale quadratically with number of nodes for full attention; however, there are methods like sampling or sparse attention to cope. In a real-world deployment covering the whole globe, we might partition the graph (e.g. analyze each country separately with linking for trade) to keep things tractable. The current model could act as a blueprint.

\ding{187} Other Impacts: $CO_2$ is one aspect; industry also affects air pollution ($SO_2$, $NOx$, $PM2.5$) and water use, jobs, etc. A holistic policy needs all that. Our model could be extended with multi-task learning to predict not just $CO_2$ but also pollution indicators, to see co-benefits or trade-offs. For example, cutting coal reduces $CO_2$ and $SO_2$ (co-benefit), but other efficiency measures might reduce $CO_2$ while increasing certain types of waste, etc. Multi-output modeling would be a future enhancement.

\section{Related Work}
\paragraph{Traditional Emission Modeling} Early approaches to understanding carbon emissions in industry relied on environmental economics and statistical frameworks. One foundational model is IPAT/STIRPAT equation, which expresses environmental impact as a product of factors like population, affluence, and technology~\cite{york2003stirpat}. Extensions of this with index decomposition analysis have been applied to dissect drivers of $CO_2$ emissions. For example, studies use STIRPAT to attribute emissions changes to economic growth vs. energy intensity~\cite{tao2024dynamics}. While useful for high-level insights, these linear models assume independent effects and often fail to capture nonlinear interactions or collinear drivers. In industrial contexts, multiple energy sources often rise and fall together, violating model assumptions and making it hard to isolate each fuel’s contribution. Researchers have noted that multicollinearity can inflate uncertainty in regression coefficients and obscure causal inference in such analyses~\cite{kim2019multicollinearity}. Various statistical remedies like principal component analysis for dimensionality reduction or ridge regression have been tried to alleviate multicollinearity~\cite{chan2022mitigating}. However, these techniques either sacrifice interpretability or only partially address correlation by shrinking coefficients without exploiting relationships between features. A more domain-specific tactic has been to group or cluster data a priori to reduce heterogeneity~\cite{zhang2023multicollinearity}. While promising, this approach still treats clusters independently and uses linear model within each, potentially missing cross-cluster interactions or nonlinear patterns. Our work builds on the intuition of clustering but learns the structure automatically via graph neural network, allowing soft clustering and joint model training.

\paragraph{Deep Learning for Emissions Forecasting} 
With the rise of machine learning, numerous studies have applied regression trees, ensembles, and neural networks to forecast energy consumption. For example, ARIMA and SVR were combined to predict industrial emissions for medium-scale data in certain regions~\cite{gao2023review}. In recent years, deep learning has become the state-of-the-art for many prediction tasks, including carbon emissions. Recurrent neural networks like LSTMs excel at capturing temporal dependencies and outperformed traditional models~\cite{waqas2024critical}. Similarly, CNN-LSTM hybrids achieved good accuracy for short-term energy structure prediction~\cite{agga2022cnn}. However, these deep models are often black-box in nature. Pure LSTMs or MLPs typically do not inherently resolve multicollinearity – if two input features move together, a neural network arbitrarily distributes weights between them, or worse, oscillates during training~\cite{ahmed2023deep}. Some researchers incorporated feature selection or attention mechanisms to tackle this, learning to focus on more informative feature among correlated set~\cite{mumuni2025automated}. Yet, standard attention alone doesn’t incorporate domain structure beyond sequence context. To mitigate this, Temporal Graph Transformer Network combines LSTM sub-networks with Graph Attention Networks to forecast carbon emissions across regions~\cite{wu2024carbon}. Likewise, Spatial-Temporal Graph Transformer that uses graph convolution for spatial dependency and attention-based transformer for time series demonstrating state-of-the-art results on emission-like data~\cite{wang2024multiscale}. Our framework is inspired by such successes, but extending them by targeting industrial sector data specifically and emphasizing interpretability for social impact.
\end{document}